%% file: main.tex
\documentclass[runningheads]{llncs}
\input{macro}

\begin{document}
\pagestyle{headings}
\mainmatter
\def\ECCVSubNumber{4452}  

\title{\TITLE} 

\titlerunning{D$^3$Net}
%
\author{
Dave Zhenyu Chen$^{1}$\index{Chen, Zhenyu} \quad Qirui Wu$^{2}$ \quad Matthias Nie{\ss}ner$^{1}$ \quad Angel X. Chang$^{2}$\index{Chang, Angel} \\
$^{1}$Technical University of Munich \qquad $^{2}$Simon Fraser University \\
\url{https://daveredrum.github.io/D3Net/} \\
}
\authorrunning{Chen et al.}
\institute{}
%

\maketitle

\input{figures/teaser}

\begin{abstract}

\input{sections/0_abstract}

\end{abstract}


\input{sections/1_intro}
\input{sections/2_related_work}
\input{sections/3_Method}
\input{sections/4_Experiments}

\input{sections/5_Conclusion}
\input{sections/7_acknowledgements}

\clearpage
%
%
{\small
\bibliographystyle{splncs04nat}
\bibliography{egbib}
}

\clearpage

\appendix

\section*{Supplementary Material}
\input{sections/6_supplemental}

\end{document}

%% file: macro.tex
\usepackage{graphicx}

\usepackage{tikz}
\usepackage{comment}
\usepackage{amsmath,amssymb} 
\usepackage{color}

\usepackage[accsupp]{axessibility}  

\usepackage{booktabs}
\usepackage{multirow}
\usepackage{caption}
\usepackage{float}
\usepackage{subcaption}
\usepackage{enumitem}
\usepackage[numbers]{natbib}
\usepackage{xspace}
\usepackage{subcaption}

\makeatletter 
\renewcommand\@biblabel[1]{#1.} 
\makeatother

\usepackage[pagebackref,breaklinks,colorlinks]{hyperref}

\newcommand{\ARCH}{D$^3$Net\xspace}
\newcommand{\TITLE}{\ARCH: A Unified Speaker-Listener Architecture for 3D Dense Captioning and Visual Grounding}

\newcommand{\Lspkmle}{L_{\text{spk-XE}}\xspace}
\newcommand{\Lspkrl}{L_{\text{spk-R}}\xspace}

\usepackage[capitalize]{cleveref}
\crefname{section}{Sec.}{Secs.}
\Crefname{section}{Section}{Sections}
\Crefname{table}{Table}{Tables}
\crefname{table}{Tab.}{Tabs.}

\def\cvprPaperID{4258} 
\def\confName{CVPR}
\def\confYear{2022}

\newcommand{\mypara}[1]{\noindent\textbf{#1}}

\newcommand{\denselist}{\itemsep 0pt\parsep=0pt\partopsep 0pt\vspace{-\topsep}}

\newcommand{\mypar}[1]{\noindent \textbf{#1}}

%% file: figures/teaser.tex
\begin{figure}
    \centering
    \includegraphics[width=0.99\textwidth]{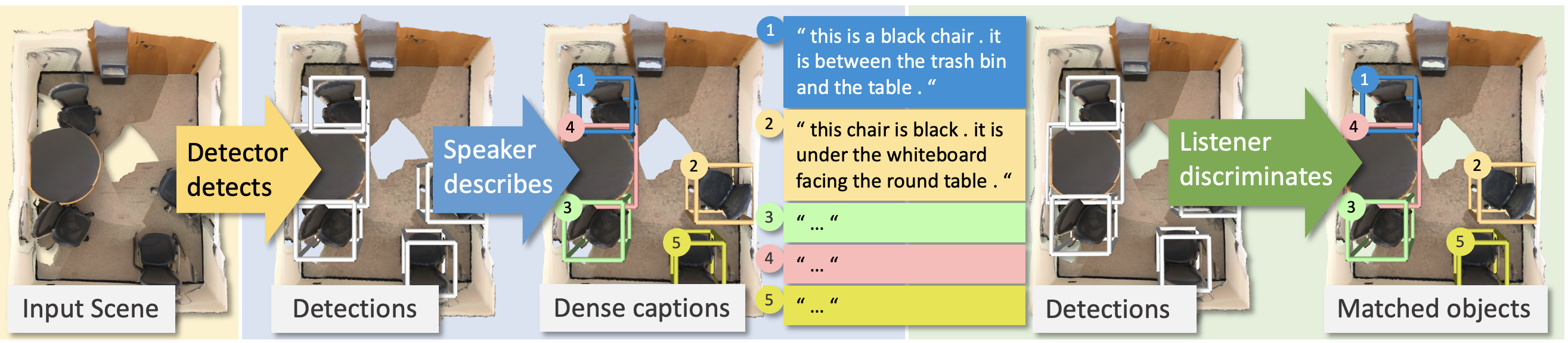}
    \caption{
    We introduce \ARCH, an end-to-end neural speaker-listener architecture that can \textbf{d}etect, \textbf{d}escribe and \textbf{d}iscriminate. \ARCH~also enables semi-supervised training on ScanNet data with partially annotated descriptions.
    }
    \label{fig:teaser}
\end{figure}

%% file: sections/0_abstract.tex
Recent work on dense captioning and visual grounding in 3D have achieved impressive results. 
Despite developments in both areas, the limited amount of available 3D vision-language data causes overfitting issues for 3D visual grounding and 3D dense captioning methods.
Also, how to discriminatively describe objects in complex 3D environments is not fully studied yet.
To address these challenges, we present \ARCH, an end-to-end neural speaker-listener architecture that can \textbf{d}etect, \textbf{d}escribe and \textbf{d}iscriminate.
Our \ARCH unifies dense captioning and visual grounding in 3D in a self-critical manner.
This self-critical property of \ARCH encourages generation of discriminative object captions and enables semi-supervised training on scan data with partially annotated descriptions.
Our method outperforms SOTA methods in both tasks on the ScanRefer dataset, surpassing the SOTA 3D dense captioning method by a significant margin.

%% file: sections/1_intro.tex
\input{figures/motivation}

\section{Introduction}
\label{sec:introduction}

Recently, there has been increasing interest in bridging 3D visual scene understanding~\citep{qi2019deep, hou20193d, hou2020revealnet, chang2017matterport3d, dai2017scannet, hua2016scenenn, song2015sun} and natural language processing~\citep{vaswani2017attention, devlin2018bert, brown2020language, liu2019roberta, yang2019xlnet}.
The task of 3D visual grounding~\citep{chen2020scanrefer, yuan2021instancerefer, zhao20213dvg} localizes 3D objects described by natural language queries.
3D dense captioning proposed by~\citet{chen2021scan2cap} is the reverse task where we generate descriptions for 3D objects in RGB-D scans.
Both tasks enable applications such as assistive robots and natural language control in AR/VR systems.


However, existing work on 3D visual grounding~\citep{chen2020scanrefer, achlioptas2020referit3d, yuan2021instancerefer, huang2021text, zhao20213dvg} and dense captioning~\citep{chen2021scan2cap, yuan2022xtrans2cap} treats the two problems as separate, with 
\emph{detect-then-dis-}\emph{criminate} or \emph{detect-then-describe} being the common strategies for tackling the two tasks.
Separating the two complementary tasks hinders holistic 3D scene understanding where the ultimate goal is to create models that can infer: 1) what are the objects; 2) how to describe each object; 3) what object is being referred to through natural language.
The disadvantages of having separated strategies are twofold.
First, the detect-then-describe strategy often struggles to describe target objects in a discriminative way. 
In Fig.~\ref{fig:motivation}, the generated descriptions from Scan2Cap~\citep{chen2021scan2cap} fail to uniquely describe the target objects, especially in scenes with several similar objects.
Second, existing 3D visual grounding methods~\citep{chen2020scanrefer, zhao20213dvg} in the detect-then-discriminate strategy suffer from severe overfitting issue, partly due to the small amount of 3D vision-language data~\citep{chen2020scanrefer, achlioptas2020referit3d} which is limited compared to counterpart 2D datasets such as MSCOCO~\citep{lin2014microsoft}.

To address these issues, we propose an end-to-end self-critical solution, \ARCH, to enable discriminability in dense caption generation and utilize the generated captions improve localization.
Relevant work in image captioning~\citep{luo2018discriminability, liu2018show} tackles similar issues where the generated captions are indiscriminative and repetitive by explicitly reinforcing discriminative caption generation with an image retrieval loss.
Inspired by this scheme, we introduce a speaker-listener strategy, where the captioning module ``speaks'' about the 3D objects, while the localization module ``listens'' and finds the targets.
Our proposed speaker-listener architecture can \textbf{d}etect, \textbf{d}escribe and \textbf{d}iscriminate, as illustrated in Fig.~\ref{fig:teaser}.
The key idea is to reinforce the speaker to generate discriminative descriptions so that the listener can better localize the described targets given those descriptions.

This approach brings another benefit.
Since the speaker-listener architecture self-critically generates and discriminates descriptions, we can train on scenes without any object descriptions.
We see further improvements in 3D dense captioning and 3D visual grounding performance when using this additional data alongside annotated scenes.
This can allow for semi-supervised training on RGB-D scans beyond the ScanNet dataset.
To summarize, our contributions are:
\begin{itemize}\denselist
\item We introduce a unified speaker-listener architecture to generate discriminative object descriptions in RGB-D scans.  Our architecture allows for a semi-supervised training scheme that can alleviate data shortage in the 3D vision-language field.
\item We study how the different components impact performance and find that having a strong detector is essential, and that by jointly optimizing the detector, speaker, and listener we can improve detection as well as 3D dense captioning and visual grounding.
\item We show that our method outperforms the state-of-the-art for both 3D dense captioning and 3D visual grounding method by a significant margin.
\end{itemize}

%% file: figures/motivation.tex

\begin{figure}[t]
    \centering
    \includegraphics[width=0.65\linewidth]{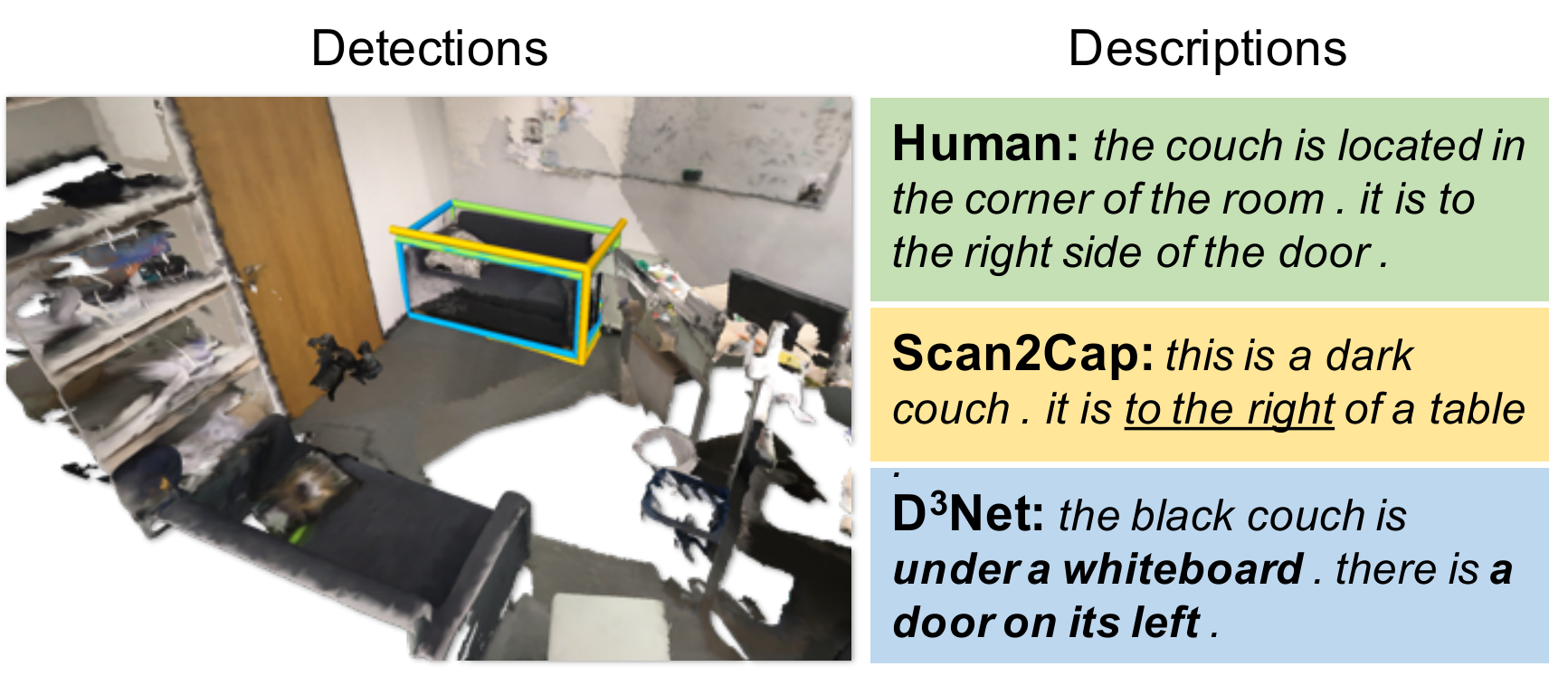}
    \caption{Prior work~\citep{chen2021scan2cap} struggle to produce discriminative object captions. Also, captions often appear to be template-based. In contrast, our \ARCH generates discriminative object captions.}
    \label{fig:motivation}
\end{figure}

%% file: sections/2_related_work.tex
\section{Related Work}
\label{sec:related}

\mypar{Vision and language in 3D.}
Recently, there has been growing interest in grounding language to 3D data~\cite{chen2018text2shape, achlioptas2019shapeglot, chen2020scanrefer, achlioptas2020referit3d, wu2021towers, roh2021languagerefer, thomason2021language}.
%
%
\citet{chen2020scanrefer} and~\citet{achlioptas2020referit3d} introduce two complementary datasets consisting of descriptions of real-world 3D objects from ScanNet~\citep{dai2017scannet} reconstructions, named ScanRefer and ReferIt3D, respectively.
ScanRefer proposes the joint task of detecting and localizing objects in a 3D scan based on a textual description, while ReferIt3D is focused on distinguishing 3D objects from the same semantic class given ground-truth bounding boxes.
\citet{yuan2021instancerefer} localize objects by decomposing input queries into fine-grained aspects, and use PointGroup~\citep{jiang2020pointgroup} as their visual backbone. However, the frozen detection backbone is not fine-tuned together with the localization module.
~\citet{zhao20213dvg} propose a transformer-based architecture with a VoteNet~\citep{qi2019deep} backbone to handle multimodal contexts during localization.
Despite the improved matching module, their work still suffers from poor quality detections due to the weak 3D detector.  We show that fine-tuning an improved 3D detector is essential to getting good predictions and good localization performance.
\citet{chen2021scan2cap} introduce the task of densely detecting and captioning objects in RGB-D scans.
Recently,~\citet{yuan2022xtrans2cap} aggregate the 2D features to point cloud to generate faithful object descriptions. 
Although their methods can effectively detect objects and generate captions w.r.t. their attributes, the quality of the bounding boxes and the discriminability of the captions are inadequate.
Our method explicitly handles the discriminability of the generated captions through a self-critical speaker-listener architecture, resulting in the state-of-the-art performance in both 3D dense captioning and 3D visual grounding tasks.

\mypar{Generating captions in images.}
Image captioning has attracted a great deal of interest~\citep{vinyals2015show, xu2015show, donahue2015long, karpathy2015deep, lu2017knowing, anderson2018bottom, jiang2018recurrent, rennie2017self, sidorov2020textcaps}. 
Recent work~\citep{luo2018discriminability, liu2018show} suggest that traditional encoder-decoder-based image captioning methods suffer from the discriminability issues.
~\citet{luo2018discriminability} propose an additional image retrieval branch to reinforce discriminative caption generation.
~\citet{liu2018show} propose a reinforcement learning method to train not only on annotated web images, but also images without any paired captions. 
In contrast to generating captions for the entire image, in the dense captioning task we densely generate captions for each detected object in the input image~\citep{johnson2016densecap, yang2017dense, kim2019dense}.
Although such methods are effective for generating captions in 2D images, directly applying such training techniques on 3D dense captioning can lead to unsatisfactory results, since the captions involve 3D geometric relationships.
In contrast, we work directly on 3D scene input dealing with object attributes as well as 3D spatial relationships.

\mypar{Grounding referential expressions in images.}
There has been tremendous progress in the task of grounding referential expressions in images, also known as visual grounding~\citep{kazemzadeh2014referitgame, plummer2015flickr30k, mao2016generation, hu2016natural, yu2018mattnet, hu2016segmentation}.
Given an image and a natural language text query as input, the target object is either localized by a bounding box~\citep{hu2016natural, yu2018mattnet}, or a segmentation mask~\citep{hu2016segmentation}.
These methods have achieved great success in the image domain.
However, they are not designed to deal with 3D geometry inputs and handle complex 3D spatial relationships.
Our proposed method directly decomposes the 3D input data with a sparse convolutional detection backbone, which produces accurate object proposals as well as semantically rich features.

\mypar{Speaker-listener models for grounding.}
The speaker-listener model is a popular architecture for pragmatic language understanding, where a line of research explores how the context and communicative goals affect the linguistics~\citep{cole1977syntax, golland2010game}.
Recent work use neural speaker-listener architectures to tackle referring expression generation~\citep{mao2016generation, yu2017joint, luo2017comprehension}, vision-language navigation~\citep{fried2018speaker}, and shape differentiation~\citep{achlioptas2019shapeglot}.
\citet{mao2016generation} construct a CNN-LSTM architecture optimized by a softmax loss to directly discriminate the generated referential expressions. 
There is no separate neural listener module compared with our method.
\citet{luo2017comprehension} and~\citet{yu2017joint} introduce a LSTM-based neural listener in the speaker-listener pipeline, but generating the referential expression is not directly supervised via the listener model, but rather trained via a proxy objective.
In contrast, our method directly optimizes the Transformer-based neural listener for the visual grounding task by discriminating the generated object captions without any proxy training objective.
Similarly, \citet{achlioptas2019shapeglot} includes a pretrained and frozen listener in the training objective, while ours enables joint end-to-end optimization for both the speaker and listener via policy gradient algorithm.
We experimentally show our method to be effective for semi-supervised learning in the two 3D vision-language tasks.

%% file: sections/3_Method.tex
\section{Method}
\label{sec:method}

\input{figures/architecture}

\ARCH has three components: a 3D object detector, the speaker (captioning) module, and the listener (localization) module.
Fig.~\ref{fig:architecture} shows the overall architecture and training flow.
The point clouds are fed into the detector to predict object proposals.
The speaker takes object proposals as input to produce captions.
To increase caption discriminability, we match these captions with object proposals via the listener.
Caption quality is measured by the CIDEr~\citep{vedantam2015CIDEr} scores and the listener loss, which are back-propagated via REINFORCE~\citep{williams1992simple} as rewards to the speaker.
Our architecture can handle scenes without ground-truth (GT) object descriptions by reinforcing the speaker with the listener loss only.

\subsection{Modules}

\mypar{Detector.}
We use PointGroup~\citep{jiang2020pointgroup} as our detector module.
PointGroup is a relatively simple model for 3D instance segmentation that achieves competitive performance on the ScanNet benchmark. 
We use ENet to augment the point clouds with multi-view features, following~\citet{dai20183dmv}.
PointGroup uses a U-Net architecture with a SparseConvNet backbone to encode point features, cluster the points, and uses ScoreNet, another U-Net structure, to score each cluster.
We take the cluster features after ScoreNet as the encoded object features.
We refer readers to the original paper~\cite{jiang2020pointgroup} for more details.
The object bounding boxes are determined by taking the minimum and maximum points in the point clusters, and are produced as final outputs of our detector module.

\mypar{Speaker.}
We base our speaker on the dense captioning method introduced by \citet{chen2021scan2cap}.
Our speaker module has two submodules: 1) a relational graph module, which is responsible for learning object-to-object spatial location relationships; 2) a context-aware attention captioning module, which attentively generates descriptive tokens with respect to the object attributes as well as the object-to-object spatial relationships.

\mypar{Listener.}
For the listener, we follow the architecture introduced by \citet{chen2020scanrefer} but replace the multi-modal fusion module with the transformer-based multi-modal fusion module of \citet{zhao20213dvg}.     
Our listener module has two submodules: 1) a language encoding module with a GRU cell; 2) a transformer-based multi-modal fusion module similar to~\citet{zhao20213dvg}, which attends to elements in the input query descriptions and the detected object proposals.
As in \citet{chen2020scanrefer}, we also incorporate a language object classifier to discriminate the semantics of the target objects in the input query descriptions.

\subsection{Training Objective}

The three modules are designed to be trained in an end-to-end fashion (see \Cref{fig:architecture}).
In this section, we describe the loss for each module, and how they are combined for the overall loss.

\mypar{Detection loss.}
We use the instance segmentation loss introduced in PointGroup \citep{jiang2020pointgroup} to train the 3D backbone.  The detection loss is composed of four parts: $L_{\text{det}}=L_{\text{sem}}+L_{\text{o\_reg}}+L_{\text{o\_dir}}+L_{\text{c\_score}}$.
$L_{\text{sem}}$ is a cross-entropy loss supervising semantic label prediction for each point.
$L_{\text{o\_reg}}$ is a $L_1$ regression loss constraining the learned point offsets belonging to the same cluster.
$L_{\text{o\_dir}}$ constrains the direction of predicted offset vectors, defined as the means of minus cosine similarities.
It helps regress precise offsets, particularly for boundary points of large-size objects, since these points are relatively far from the instance centroids.
$L_{\text{c\_score}}$ is another binary cross-entropy loss supervising the predicted objectness scores.

\mypar{Listener loss.}
The listener loss is composed of a localization loss $L_{\text{loc}}$ and a language-based object classification loss $L_{\text{lobjcls}}$.
To obtain the localization loss $L_{\text{loc}}$, we first require a target bounding box.
We use the detected bounding box with the highest IoU with the GT bounding box as the target bounding box.
Then, a cross-entropy loss $L_{\text{loc}}$ is applied to supervise the matching score prediction.
In the end-to-end training scenario, the detected bounding boxes associated with the generated descriptions from the speaker are treated as the target bounding boxes.
The language object classification loss is a cross-entropy loss $L_{\text{lobjcls}}$ to supervise the classification based on the input description.
The target classes are consistent with the ScanNet 18 classes, excluding structural objects such as ``floor'' and ``wall''.

\mypar{Speaker loss using MLE training objective.}
The speaker loss is a standard captioning loss from maximum likelihood estimation (MLE).
During training, provided with a pair of GT bounding box and the associated GT description, we optimize the description associated with the predicted bounding box which has the highest IoU score with the current GT bounding box.
We first treat the description generation task as a sequence prediction task, factorized as: $\Lspkmle(\theta) = -\sum_{t=1}^T\text{log}p(\hat{c_t}|\hat{c_1},...,\hat{c_{t-1}};I,\theta)$, where $\hat{c_t}$ denotes the generated token at step $t$; $I$ and $\theta$ represent the visual signal and model parameter, respectively.
The token $\hat{c_t}$ is sampled from the probability distribution over the pre-defined vocabulary.
The generation process is performed by greedy decoding or beam search in an autoregressive manner, and we use the argmax function to sample each token.

\mypar{Joint loss using REINFORCE training objective.}
We use REINFORCE to train the detector-speaker-listener jointly.
We first describe the enhanced speaker-loss, $\Lspkrl$ that is trained using reinforcement learning to produce discriminative captions.
We then describe the overall loss used in end-to-end training.
Following prior work~\citep{luo2018discriminability, liu2018show, ranzato2015sequence, hendricks2016generating, yu2017joint, rennie2017self}, generating descriptions is treated as a reinforcement learning task.
In the setting of reinforcement learning, the speaker module is treated as the ``agent'', while the previously generated words and the input visual signal $I$ are the ``environment''.
At step $t$, generating word $\hat{c_t}$ by the speaker module is deemed as the ``action'' taken with the policy $p_\theta$, which is defined by the speaker module parameters $\theta$.
Specifically, with the generated description $\hat{C}=\{c_1, ..., c_T\}$, the objective is to maximize the reward function $R(\hat{C},I)$.
We apply the ``REINFORCE with baseline'' algorithm following~\citet{rennie2017self} to reduce the variance of this loss function, where a baseline reward $R(C^*,I)$ of the description $C^*$ independent of $\hat{C}$ is introduced.
We apply beam search to sample descriptions and choose the greedily decoded descriptions as the baseline.
The simplified policy gradient is:
\begin{equation}
    \Lspkrl(\theta) \approx -(R(\hat{C},I)-R(C^*,I)) \sum_{t=1}^T\text{log}p(\hat{c_t}|I,\theta)
\end{equation}

\mypar{Rewards.}
As the word-level sampling through the argmax function is non-differentiable, the subsequent listener loss cannot be directly back-propagated through the speaker module.
A workaround is to use the gumbel softmax re-parametrization trick~\citep{jang2016categorical}.
Following the training scheme of \citet{liu2018show} and \citet{luo2018discriminability}, the listener loss can be inserted into the REINFORCE reward function to increase the discriminability of generated referential descriptions. Specifically, given the localization loss $L_{\text{loc}}$ and the language object classification loss $L_{\text{lobjcls}}$, the reward function $R(\hat{C})$ is the weighted sum of the CIDEr score of the sampled description and the listener-related losses:
\begin{equation}
    R(\hat{C},I) = R^{\text{CIDEr}}(\hat{C}, I) - \alpha [L_{\text{loc}}(\hat{C}) + \beta L_{\text{lobjcls}}(\hat{C})]
\end{equation}
where $\alpha$ and $\beta$ are the weights balancing the CIDEr reward and the listener rewards.
We empirically set them to 0.1 and 1 in our experiments, respectively.
To stabilize the training, the reward related to the baseline description $R(C^*)$ should be formulated analogously.
Note that there should be no gradient calculation and back-propagation for the baseline $C^*$.
For scenes with no GT descriptions provided, the CIDEr reward is cancelled in the reward function, which in this case becomes $R(\hat{C},I) = - \alpha [L_{\text{loc}}(\hat{C}) + \beta L_{\text{lobjcls}}(\hat{C})]$.

\mypar{Relative orientation loss.}
Following~\citet{chen2021scan2cap}, we adopt the relative orientation loss on the message passing module as a proxy loss.
The object-to-object relative orientations ranging from $0^{\circ}$ to $180^{\circ}$ are discretized into 6 classes.
We apply a simple cross-entropy loss $L_{\text{ori}}$ to supervise the relative orientation predictions.

\mypar{Overall loss.}
We combine loss terms in our end-to-end joint training objective as: $L=L_{\text{det}}+\Lspkrl+0.3L_{\text{ori}}$. 

\subsection{Training}
We use a stage-wise training strategy for stable training. We first pretrain the detector backbone on all training scans in ScanNet via the detector loss $L_\text{det}$. We then train the dense captioning pipeline with the pretrained detector and a newly initialized speaker end-to-end via the detector loss and the speaker MLE loss $L_\text{spk-XE}$. After the speaker MLE loss converges, we train the visual grounding pipeline with the fine-tuned frozen detector and the listener via the listener loss $L_\text{loc}$. Finally, we fine-tune the entire speaker-listener architecture with the overall loss $L$.

\subsection{Inference}

During inference, we use the detector and the speaker to do 3D dense captioning and the listener to do visual grounding.
The detector first produces object proposals, and the speaker generates a description for each object proposal.
We take the minimum and maximum coordinates in the predicted object instance masks to construct the bounding boxes.
For the object proposals that are assigned to the same ground truth, we keep only the one with the highest IoU with the GT bounding box.
When evaluating the detector itself, the non-maximum suppression is applied.

%% file: figures/architecture.tex
\begin{figure*}[t]
    \centering
    \includegraphics[width=\textwidth]{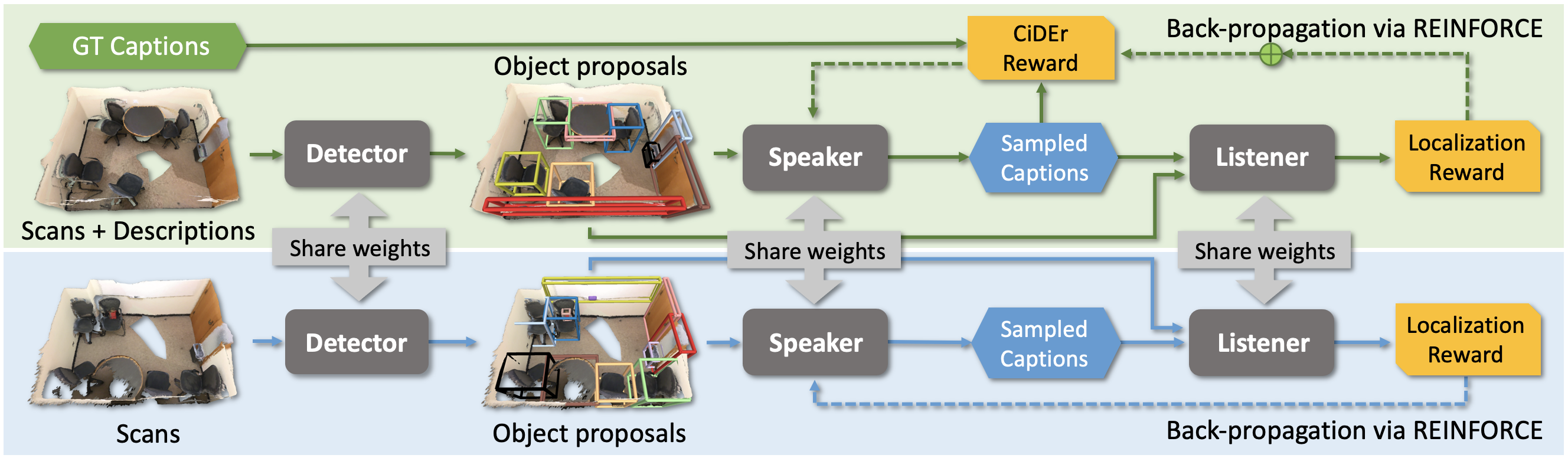}
    \caption{\ARCH~architecture. We input point clouds into the \emph{detector} to predict object proposals. Then, those proposals are fed into the speaker to generate captions that \emph{describes} each object. To \emph{discriminate} the object described by each caption, the listener matches the generated captions with object proposals. The captioning and localization results are back-propagated via REINFORCE~\citep{williams1992simple} as rewards through the dashed lines. \ARCH~also enables end-to-end training on point clouds with no GT object descriptions (bottom blue block).}
    \label{fig:architecture}
\end{figure*}

%% file: sections/4_Experiments.tex
\section{Experiments}
\label{sec:experiments}

\subsection{Dataset}

We use the ScanRefer~\citep{chen2020scanrefer} dataset consisting of around 51k descriptions for over 11k objects in 800 ScanNet~\citep{dai2017scannet} scans.
The descriptions include information about the appearance of the objects, as well as the object-to-object spatial relationships.
We follow the official split from the ScanRefer benchmark for training and validation.
We report our visual grounding results on the validation split and benchmark results on the hidden test set\footnote{\url{http://kaldir.vc.in.tum.de/scanrefer_benchmark}}.
Our dense captioning results are on the validation split due to the lack of the test grounding truth.
We also conduct experiments on the ReferIt3D dataset~\citep{achlioptas2020referit3d} (please see the supplemental).

\subsection{Semi-supervised Training with Extra Data}

As the scans in ScanRefer dataset are only a subset of scans in ScanNet, we extend the training set by including all re-scans of the same scenes for semi-supervised training.  Unlike the scans in ScanRefer, these re-scans do not have per object descriptions.
We can control how much extra data to use by randomly sampling (with replacement) from the set of re-scans.  We experiment with augmenting our data with $0.1$ to $1$ times the amount of annotated data as extra data. 
During training, we randomly select detected objects in the sampled extra scans for subsequent dense captioning and visual grounding.
For the complete `extra' scenario, we use a comparable amount (1x) of extra data as the annotated data in ScanRefer.

\subsection{Implementation Details}

We implement the PointGroup backbone using the Minkowski Engine~\citep{choy20194d} (see supplement).
For the backbone, we train using Adam~\citep{kingma2014adam} with a learning rate of 2e-3, on the ScanNet train split with batch size 4 for 140k iterations, until convergence.
For data augmentation, we follow \citet{jiang2020pointgroup}, randomly applying jitter, mirroring about the YZ-plane, and rotation about the Z axis (up-axis) to each point cloud scene.
We then use the Adam optimizer with learning rate 1e-3 to train the detector and the listener on the ScanRefer dataset with batch size 4 for 60k iterations, until convergence.
Each scan is paired with 8 descriptions (i.e. 4 scans and 32 descriptions per batch iteration).
Then, we combine the trained detector with the newly initialized speaker on the ScanRefer dataset for the 3D dense captioning task, where the weights of the detector are frozen.
We again use Adam with learning rate 1e-3, with the training process converging within 14k iterations.
All our experiments are conducted on a RTX 3090, and all neural modules are implemented using PyTorch~\citep{NEURIPS2019_9015}.

\subsection{Quantitative Results}

\input{tables/captioning}

\input{tables/grounding}

\mypara{3D dense captioning and detection}
Tab.~\ref{tab:captioning} compares our 3D dense captioning and object detection results against the baseline methods Scan2Cap~\citep{chen2021scan2cap} and X-Trans2Cap~\citep{yuan2022xtrans2cap}.
Leveraging the improved PointGroup based detector, our speaker model trained with the conventional MLE objective (Ours (MLE)) outperforms Scan2Cap and X-Trans2Cap by a large margin in all metrics.
As expected, training with the CIDEr reward (Ours (CIDEr)) significantly improves the CIDEr score.  We note that other captioning metrics are also improved, but the detection mAP@0.5 remains similar.
Training with object localization reward (Ours (CIDEr+loc.)) improves both captioning and detection further due to the improved discriminability during description generation.  Note that if we use a frozen pretrained listener (Ours (CIDEr+fixed loc.)), the improvement is not as significant as when we allow the listener weights to be fine-tuned (Ours (CIDEr+loc.)).
Our full model with the full listener reward incorporates an additional language object classification loss (Ours (CIDEr+loc.+lobjcls.)) and further improves the performance for both tasks.

\paragraph{Does additional data help?} As our method allow for training the listener with scans without language data, we investigate the effectiveness of training with additional ScanNet data that have not been annotated with descriptions.  We vary the amount of extra scan data (without descriptions) from 0.1x to 1x of fully annotated data and train our full model with CIDEr and full listener reward (loc.+lobjcls.).  
Our results (last three rows of Tab.~\ref{tab:captioning}), show that our semi-supervised training strategy can leverage the extra data to improve both dense captioning and object detection.  



\mypara{3D visual grounding}
Tab.~\ref{tab:grounding} compares our results against prior 3D visual grounding methods ScanRefer~\citep{chen2020scanrefer}, TGNN~\citep{huang2021text}, InstanceRefer~\citep{yuan2021instancerefer} and 3DVG-Transformer~\citep{zhao20213dvg}, and 3DVG-Trans+, an unpublished extension. 
Our method trained only with the detection loss and the listener loss (``Ours w/o fine-tuning''), i.e. without the speaker-listener setting, outperforms all the previous methods in the ``Unique'' and ``Overall'' scenarios.  We find the improved fusion module together with the improved detector is sufficient to outperform 3DVG-Trans.  Due to the improved detector, our method can distinguish objects in the ``Unique'' case, where the semantic labels play an important role.
Meanwhile, 3DVG-Trans~\citep{zhao20213dvg} still outperforms our base listener when discriminating objects from the same class (``Multiple'' case).
Our end-to-end speaker-listener (last row) outperforms all previous method including 3DVG-Trans.

\input{figures/captioning}

\input{figures/grounding}

\subsection{Qualitative Analysis}

\mypara{3D dense captioning}
Fig.~\ref{fig:captioning} compares our results with object captions from Scan2Cap~\citep{chen2021scan2cap}. Descriptions generated by Scan2Cap cannot uniquely identify the target object in the input scenes (see the yellow block on the bottom right).
Also, Scan2Cap produces inaccurate object bounding boxes, which affects the quality of object captions (see the yellow block on the top left).
Compared to captions from Scan2Cap, our method produces more discriminative object captions that specifies more spatial relations (see bolded phrases in the blue blocks).

\mypara{3D visual grounding}
Fig.~\ref{fig:grounding} compares our results with 3DVG-Transformer~\citep{zhao20213dvg}.
Though 3DVG-Transformer is able to pick the correct object, it suffers from poor object detections and is constrained by the performance of the VoteNet-based detection backbone (see the first column).
Our method is capable of selecting the queried objects while also predicting more accurate object bounding boxes.

\input{tables/different_detector}

\input{tables/discrimination}

\input{tables/manual_analysis}

\subsection{Analysis and Ablation Studies}
\mypar{Does better detection backbone help?} 
From Tab.~\ref{tab:captioning}, we see that using a better detector can significant improve performance.  We further examine the effect of using different detection backbones (VoteNet and PointGroup) compared to GT bounding boxes in Tab.~\ref{tab:different_detection}. 
For each detection backbone, we use four variants of our method: the models trained without the joint speaker-listener architecture, and the speaker-listener architecture trained with CIDEr reward, listener reward and extra ScanNet data. The results with GT boxes show the effectiveness of our speaker-listener architecture, when detections are perfect. The large improvement from VoteNet~\citep{qi2019deep} to PointGroup~\citep{jiang2020pointgroup} show the importance of a better detection backbone. The gap between GT and VoteNet/PointGroup shows there is room for further improvement.

\mypar{Are the generated descriptions more discriminative?}
To check whether the speaker-listener architecture generates more discriminative descriptions, we conduct an automatic evaluation via a reverse task.  In this task, we feed the generated descriptions and GT bounding boxes into a pretrained neural listener model similar to \citet{zhao20213dvg}.
The predicted visual grounding results are evaluated in the same way as in our 3D visual grounding experiments.
Higher grounding accuracy indicates better discrimination, especially in the ``Multiple'' case.  Results (Tab.~\ref{tab:discrimination}) show that our speaker-listener architecture generates more discriminative descriptions compared to Scan2Cap~\citep{chen2021scan2cap}.
The discrimination is further improved when training with extra ScanNet data.
To disentangle the affect of imperfectly predicted bounding boxes, we also train and evaluate our method with GT boxes  (see last two rows in Tab.~\ref{tab:discrimination}).
We see that our semi-supervised speaker-listener architecture generates more discriminative descriptions.

\mypara{Does the listener help with captioning?}
The third to the sixth rows in Tab.~\ref{tab:captioning} measure the benefit of training the speaker together with the listener (Ours (CIDEr+loc.) and Ours (CIDEr+loc.+lobjcls.)) rather than training the speaker alone (Ours (CIDEr)).
Training with the listener improves all captioning metrics.
Also, training jointly with an unfrozen listener (Ours (CIDEr+loc.) leads to a better performance when compared with the variant with a pretrained and frozen listener (Ours (CIDEr+fixed loc.), which is similar to~\citet{achlioptas2019shapeglot}.
Additionally, as the detector is not only fine-tuned with the speaker but also with the listener, the additional supervision from the listener helps with the detection performance as well. 

To analyze the quality of the generated object captions, we asked $5$ students to perform a fine-grained manual analysis of the captions. Each student was presented with a batch of $100$ randomly selected object captions with associated objects highlighted in the 3D scene. The student are then asked to indicate if the respective aspects were included and correctly described. 
The manual analysis results in Tab.~\ref{tab:manual_analysis} shows that our method generates more accurate descriptions compared to Scan2Cap.
In particular, training with the listener and extra ScanNet data produces more accurate spatial relations in the descriptions.
The results of fine-grained manual analysis complements the automatic captioning evaluation metric. While metrics such as CIDEr captures the overall similarity of the generated sentences against the references, the accuracies in Tab.~\ref{tab:manual_analysis} measures the correctness of the decomposed visual attributes.

\mypar{Does the speaker help with grounding?}
Tab.~\ref{tab:grounding} compares grounding results between a pretrained listener (Ours w/o fine-tuning) and a fine-tuned speaker-listener model (Ours).
Although the grounding performance drops in the ``Unique'' subset, the improvements in ``Multiple'' suggests better discriminability in tougher and ambiguous scenarios. 


%% file: tables/captioning.tex
\begin{table}[!t]
    \centering
    \caption{Quantitative results on 3D dense captioning and object detection. As in~\citet{chen2021scan2cap}, we average the conventional captioning evaluation metrics with the percentage of the predicted bounding boxes whose IoU with the GTs are higher than 0.5. Our speaker model outperforms the baseline Scan2Cap without training via REINFORCE, while training with CIDEr reward further boosts the dense captioning performance. We also showcase the effectiveness of training with additional scans with no description annotations. Our speaker-listener architecture trained with 1x extra data achieves the best performance.}
    \resizebox{0.75\linewidth}{!}{
        \begin{tabular}{l|cccc|c}
            \toprule
            & C@0.5IoU & B-4@0.5IoU & M@0.5IoU & R@0.5IoU & mAP@0.5 \\
            \midrule
            Scan2Cap~\citep{chen2021scan2cap} & 39.08 & 23.32 & 21.97 & 44.78 & 32.21 \\
            X-Trans2Cap~\citep{yuan2022xtrans2cap} & 43.87 & 25.05 & 22.46 & 44.97 & 35.31 \\
            \midrule
            Ours (MLE) & 46.07 & 30.29 & 24.35 & 51.67 & 50.93 \\
            Ours (CIDEr) & 57.88 & 32.64 & 24.86 & 52.26 & 51.01\\
            Ours (CIDEr+fixed loc.) & 58.93 & 33.36 & 25.12 & 52.62 & 51.04\\
            Ours (CIDEr+loc.) & 61.30 & 34.50 & 25.25 & 52.80 & 52.07\\
            Ours (CIDEr+loc.+lobjcls.) & 61.50 & 35.05 & 25.48 & 53.31 & 52.58\\
            \midrule
            Ours (w/ 0.1x extra data) & 61.91 & 35.03 & 25.38 & 53.25 & 52.64\\
            Ours (w/ 0.5x extra data) & 62.36 & 35.54 & 25.43 & 53.67 & 53.17\\
            Ours (w/ 1x extra data) & \textbf{62.64} & \textbf{35.68} & \textbf{25.72} & \textbf{53.90} & \textbf{53.95} \\
            \bottomrule
        \end{tabular}
    }
    
    \label{tab:captioning}
\end{table}

%% file: tables/grounding.tex
\begin{table}[!t]
    \centering
    \caption{Quantitative results on 3D visual grounding. We adapt the evaluation setting as in~\citet{chen2020scanrefer}. ``Unique'' means there is only one object belongs to a specific class in the scene, while ``multiple'' represents the cases where more than one object from a specific class can be found in the scene. Clearly, our base visual grounding network outperforms all baselines even before being put into the speaker-listener architecture. After the speaker-listener fine-tuning, our method achieves the state-of-the-art performance on the ScanRefer validation set and the public benchmark. Note that 3DVG-Trans+ is an unpublished extension of 3DVG-Trans~\citep{zhao20213dvg} which appears only on the public benchmark.}
    \resizebox{0.65\linewidth}{!}{
        \begin{tabular}{l|ccc|ccc}
            \toprule
            & \multicolumn{3}{c}{Val Acc@0.5IoU} & \multicolumn{3}{c}{Test Acc@0.5IoU} \\ 
            \cmidrule(lr){2-4} \cmidrule(lr){5-7}
            & Unique & Multiple & Overall & Unique & Multiple & Overall \\
            \midrule
            ScanRefer~\citep{chen2020scanrefer} & 53.51 & 21.11 & 27.40 & 43.53 & 20.97 & 26.03 \\
            TGNN~\citep{huang2021text} & 56.80 & 23.18 & 29.70 & 58.90 & 25.30 & 32.80\\
            InstanceRefer~\citep{yuan2021instancerefer} & 66.83 & 24.77 & 32.93 & 66.69 & 26.88 & 35.80 \\
            3DVG-Trans~\citep{zhao20213dvg} & 60.64 & 28.42 & 34.67 & 55.15 & 29.33 & 35.12\\
            3DVG-Trans+~\citep{zhao20213dvg} & - & - & - & 57.87 & \textbf{31.02} & 37.04\\
            \midrule
            Ours (w/o fine-tuning) & 70.35 & 27.11 & 35.58 & 65.79 & 27.26 & 35.90 \\
            Ours & \textbf{72.04} & \textbf{30.05} & \textbf{37.87} & \textbf{68.43} & 30.74 & \textbf{39.19} \\
            \bottomrule
        \end{tabular}
    }
    
    \label{tab:grounding}
\end{table}

%% file: figures/captioning.tex
\begin{figure*}[t]
    \centering
    \includegraphics[width=0.99\textwidth]{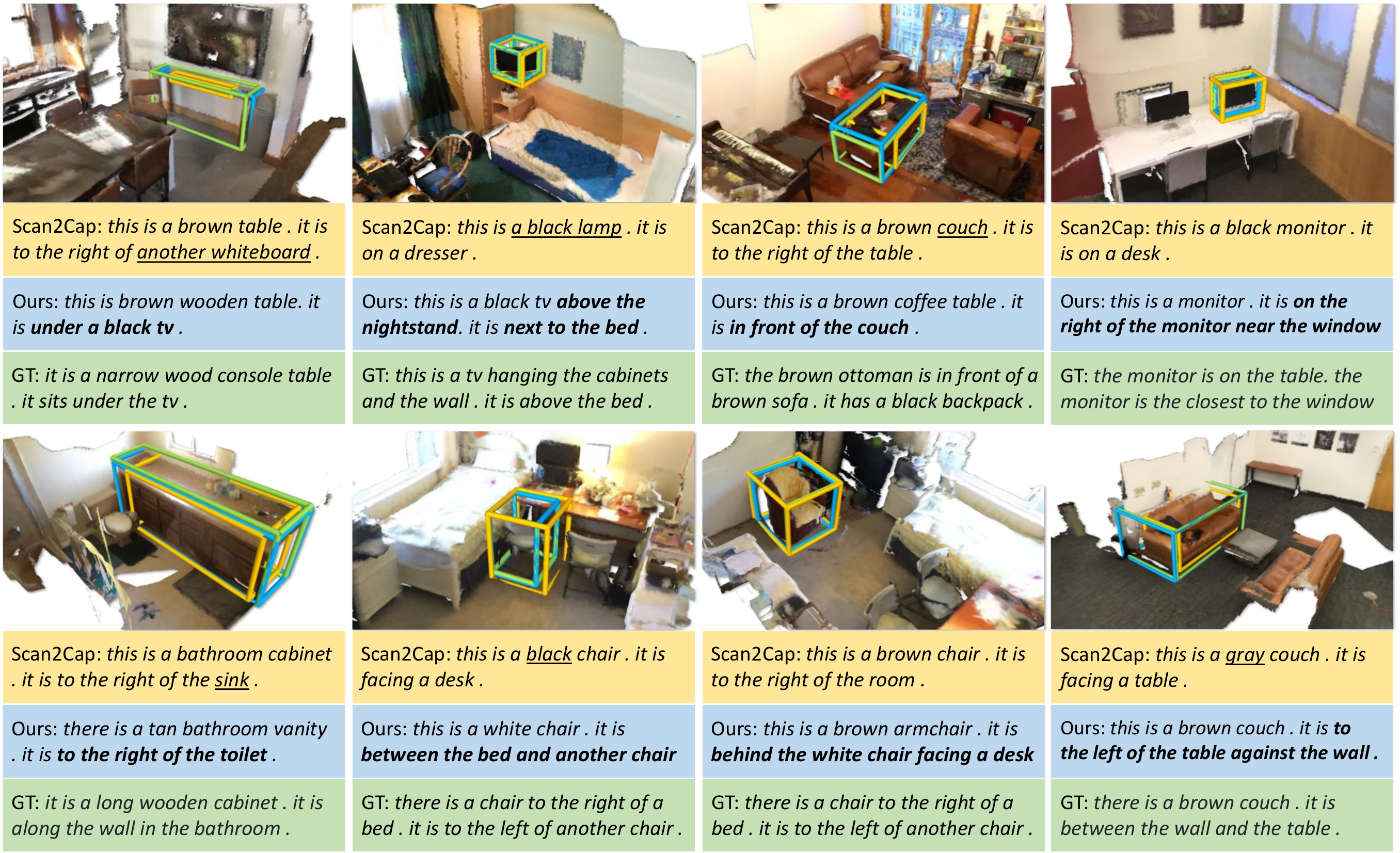}
    \caption{Qualitative results in 3D dense captioning task from Scan2Cap~\citep{chen2021scan2cap} and our method. We underline the inaccurate words and mark the spatially discriminative phrases in bold. Our method qualitatively outperforms Scan2Cap in producing better object bounding boxes and more discriminative descriptions.}
    \label{fig:captioning}
\end{figure*}

%% file: figures/grounding.tex
\begin{figure*}[t]
    \centering
    \includegraphics[width=0.99\textwidth]{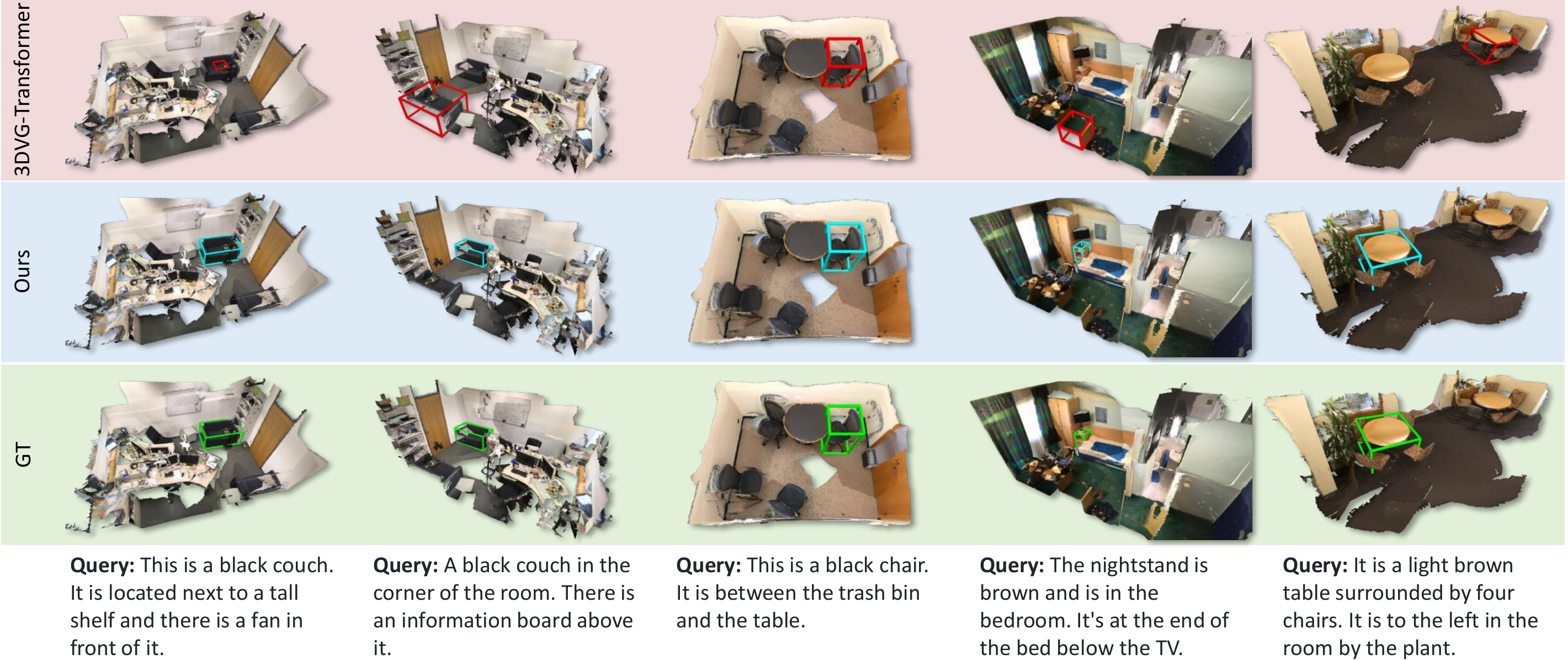}
    \caption{3D visual grounding results using 3DVG-Transformer~\citep{zhao20213dvg} and our method. 3DVG-Transformer fails to accurately predict object bounding boxes, while our method produces accurate bounding boxes and correctly distinguishes target objects from distractors.}
    \label{fig:grounding}
\end{figure*}

%% file: tables/different_detector.tex
\begin{table*}[t]
    \centering
    \caption{Quantitative results on object detection, dense captioning and visual grounding in RGB-D scans. We train our method using different detection backbones as well as the ground truth bounding boxes. Our method trained with CIDEr and listener reward as well as the additional data outperforms the pretrained speaker and listener models.}
    \resizebox{\textwidth}{!}{
        \begin{tabular}{l|c|c|cccc|ccc}
            \toprule
             & & & & & & & Unique & Multiple & Overall \\
            Method & Detection & mAP@0.5 & C@0.5IoU & B-4@0.5IoU & M@0.5IoU & R@0.5IoU & Acc@0.5IoU & Acc@0.5IoU & Acc@0.5IoU \\
            \midrule
            Ours (MLE) & GT & 100.00 & 71.41 & 42.95 & 29.67 & 64.93 & 88.45 & 36.46 & 46.03 \\
            Ours (CIDEr) & GT & 100.00 & 94.80 & 47.92 & 30.80 & \textbf{66.34} & - & - & - \\
            Ours (CIDEr+lis.) & GT & 100.00 & 95.62 & 47.65 & \textbf{30.93} & 66.31 & 89.76 & 36.85 & 47.14 \\
            Ours (CIDEr+lis.+extra) & GT & 100.00 & \textbf{96.31} & \textbf{48.20}  & 30.80 & 66.10 & \textbf{89.86} & \textbf{40.66} & \textbf{48.17} \\
            \midrule
            Ours (MLE)  & VoteNet & 32.21 & 39.08 & 23.32 & 21.97 & \textbf{44.78} & 56.41 & 21.11 & 27.95 \\
            Ours (CIDEr) & VoteNet & 37.66 & 46.88 & 25.96 & 22.10 & 44.69 & - & - & - \\
            Ours (CIDEr+lis.) & VoteNet & 38.03 & 47.32 & 24.76 & 21.66 & 43.62 & 57.90 & 20.73 & 28.03 \\
            Ours (CIDEr+lis.+extra) & VoteNet & \textbf{38.82} & \textbf{48.38} & \textbf{26.09} & \textbf{22.15} & 44.74 & \textbf{58.40} & \textbf{21.66} & \textbf{29.25} \\
            \midrule
            Ours (MLE) & PointGroup & 47.19 & 46.07 & 30.29 & 24.35 & 51.67 & 70.35 & 27.11 & 35.58 \\
            Ours (CIDEr) & PointGroup & 52.44 & 57.88 & 32.64 & 24.86 & 52.26 & - & - & - \\
            Ours (CIDEr+lis.) & PointGroup & 52.58 & 61.50 & 35.05 & 25.48 & 53.31 & 71.04 & 27.40 & 35.62 \\
            Ours (CIDEr+lis.+extra) & PointGroup & \textbf{53.95} & \textbf{62.64} & \textbf{35.68} & \textbf{25.72} & \textbf{53.90} & \textbf{72.04} & \textbf{30.05} & \textbf{37.87} \\
            \bottomrule
        \end{tabular}
    }
    \label{tab:different_detection}
\end{table*}

%% file: tables/discrimination.tex
\begin{table}[!t]
    \centering
    \caption{We automatically evaluate the discriminability of the generated object descriptions. A pretrained neural listener similar to~\citet{zhao20213dvg} is fed with the GT object features and the descriptions generated by Scan2Cap~\citep{chen2021scan2cap} as well as our method. Higher grounding accuracy indicates better discriminability, especially in the ``multiple'' case. To alleviate noisy detections, the evaluation results on the descriptions generated from the GT object features are also presented. Our method generates more discriminative descriptions compared to Scan2Cap.}
    \resizebox{0.8\linewidth}{!}{
        \begin{tabular}{l|c|ccc}
            \toprule
            & & Unique & Multiple & Overall \\
            & detection & Acc@0.5IoU & Acc@0.5IoU & Acc@0.5IoU \\
            \midrule
            Scan2Cap~\citep{chen2021scan2cap} & VN~\citep{qi2019deep} & 80.52 & 29.95 & 39.08 \\
            \midrule
            Ours (w/ CIDEr \& lis.) & PG~\citep{jiang2020pointgroup} & 81.16 & 30.22 & 41.62 \\
            Ours (w/ CIDEr \& lis. \& extra) & PG~\citep{jiang2020pointgroup} & \textbf{81.27} & \textbf{30.33} & \textbf{41.73} \\
            \midrule
            Ours (w/ CIDEr \& lis.) & GT & 89.76 & 38.53 & 48.07 \\
            Ours (w/ CIDEr \& lis. \& extra)& GT & \textbf{90.29} & \textbf{40.66} & \textbf{49.71} \\
            \bottomrule
        \end{tabular}
    }
    
    \label{tab:discrimination}
\end{table}

%% file: tables/manual_analysis.tex
\begin{table}[t]
    \centering
    \caption{Manual analysis of captions generated by Scan2Cap~\citep{chen2021scan2cap} and variants of our method. We measure accuracy in three different aspects: object categories, appearance attributes and spatial relations. Our method generates more accurate descriptions in all aspects, especially for describing spatial relations.}
    \resizebox{0.7\linewidth}{!}{
        \begin{tabular}{l|ccc}
            \toprule
             & Acc (Category) & Acc (Attribute) & Acc (Relation) \\
            \midrule
            Scan2Cap~\cite{chen2021scan2cap} & 84.10 & 64.21 & 69.00  \\
            \midrule
            Ours (MLE) & 88.00 (+3.84) & 74.73 (+10.53) & 69.00 (+0.00) \\
            Ours (CIDEr) & 88.89 (+4.73) & 75.00 (+10.79) & 68.00 (-1.00) \\
            Ours (CIDEr+lis.) & 90.91 (+6.75) & 77.38 (+13.17) & 75.00 (+6.00) \\
            Ours (CIDEr+lis.+extra) & 92.93 \textbf{(+8.77)} & 80.95 \textbf{(+16.74)} & 78.57 \textbf{(+9.57)} \\
            \bottomrule
        \end{tabular}
    }
    
    \label{tab:manual_analysis}
    \vspace{-3mm}
\end{table}

%% file: sections/5_Conclusion.tex
\section{Conclusion}
\label{sec:conclusion}

We present \ARCH, an end-to-end speaker-listener architecture that can \textbf{d}etect, \textbf{d}escribe and \textbf{d}iscriminate.
Specifically, the speaker iteratively generates descriptive tokens given the object proposals detected by the detector, while the listener discriminates the object proposals in the scene with the generated captions.
%
%
The self-discriminative property of \ARCH~also enables semi-supervised training on ScanNet data without the annotated descriptions.
Our method outperforms the previous SOTA methods in both tasks on ScanRefer, surpassing the previous SOTA 3D dense captioning method by a significant margin.
%
Our architecture can serve as an initial step towards leveraging unannotated 3D data for language and 3D vision. 
Overall, we hope that our work will encourage more future research in 3D vision and language.

%% file: sections/7_acknowledgements.tex
\section*{Acknowledgements}

This work is funded by Google (AugmentedPerception), the ERC Starting Grant Scan2CAD (804724), and a Google Faculty Award. We would also like to thank the support of the TUM-IAS Rudolf M{\"o}{\ss}bauer and Hans Fischer Fellowships (Focus Group Visual Computing), as well as the the German Research Foundation (DFG) under the Grant \textit{Making Machine Learning on Static and Dynamic 3D Data Practical}.  This work is also supported in part by the Canada CIFAR AI Chair program and an NSERC Discovery Grant.

%% file: sections/6_supplemental.tex

In this supplementary material, we provide results on the ReferIt3D dataset in Sec.~\ref{sec:referit3d}.
To showcase the effectiveness of our speaker-listener architecture, we provide additional results on extra ScanNet~\citep{dai2017scannet} data in Sec.~\ref{sec:extra}.
We also include details about our PointGroup implementation as well as the detection and segmentation results in Sec.~\ref{sec:pointgroup_details} and Sec.~\ref{sec:pointgroup_vis}, respectively.

\section{Experiments on ReferIt3D}
\label{sec:referit3d}

\subsection{Quantitative Results}

We conduct additional experiments on the ReferIt3D Nr3D dataset~\citep{achlioptas2020referit3d}.
It contains about 33k free-form object descriptions annotated by human experts for training and 8k for validation.
We report our results on the validation split since there is no test set.

\subsubsection{3D dense captioning}

We compare our 3D dense captioning and object detection results against the baseline Scan2Cap~\citep{chen2021scan2cap} in Tab.~\ref{tab:captioning_referit3d}. Our method trained with the speaker MLE loss (marked ``Ours (MLE)'') outperforms Scan2Cap by a big margin, leveraging the improved object detection backbone. After training with the CIDEr reward (marked ``Ours (CIDEr)''), our dense captioning results are further boosted. Training with the listener loss as the additional reward (marked ``Ours (CIDEr+lis.)'') further improves our results due to the explicit reinforcement of the discriminability of generated object descriptions. Here, our object detection mAP is also improved due to the end-to-end joint fine-tuning of our speaker-listener architecture. We showcase the effectiveness of training with extra ScanNet data in the last row in Tab.~\ref{tab:captioning_referit3d}, where 3D dense captioning and object detection results are improved simultaneously.

\begin{table}[!h]
    \centering
    \caption{Quantitative results on 3D dense captioning and object detection on ReferIt3D Nr3D dataset~\citep{achlioptas2020referit3d}. We average the conventional captioning evaluation metrics with the percentage of the predicted bounding boxes whose IoU with the GTs are higher than 0.5. Our method outperforms the baseline Scan2Cap~\citep{chen2021scan2cap} by a significant margin. We showcase the effectiveness of our speaker-listener architecture trained with partially annotated ScanNet data, where it achieves the best performance in all metrics.}
    \resizebox{0.9\linewidth}{!}{
        \begin{tabular}{l|cccc|c}
            \toprule
            & C@0.5IoU & B-4@0.5IoU & M@0.5IoU & R@0.5IoU & mAP@0.5 \\
            \midrule
            Scan2Cap~\citep{chen2021scan2cap} & 22.38 & 13.87 & 20.44 & 47.96 & 33.21 \\
            \midrule
            Ours (MLE) & 33.85 & 20.70 & 23.13 & 53.38 & 49.71 \\
            Ours (CIDEr) & 36.79 & 21.12 & 23.91 & 53.83 & 50.89 \\
            Ours (CIDEr+lis.) & 37.35 & 21.40 & 24.10 & 54.14 & 51.58 \\
            Ours (CIDEr+lis.+extra) & \textbf{38.42} & \textbf{22.22} & \textbf{24.74} & \textbf{54.37} & \textbf{52.69} \\
            \bottomrule
        \end{tabular}
    }
    
    \label{tab:captioning_referit3d}
\end{table}

\subsubsection{3D visual grounding}

\begin{table}[!h]
    \centering
    \caption{Quantitative results on 3D visual grounding on ReferIt3D Nr3D dataset~\citep{achlioptas2020referit3d}. We adapt
    the evaluation setting as in~\citet{chen2020scanrefer} to be consistent with the main paper. We report results on ``Multiple'' and ``Overall'', as there is no case in ReferIt3D that is ``Unique''. Our base visual grounding network outperforms the baseline methods. Results are further improved after the joint fine-tuning with the speaker-listener architecture. Speaker-listener fine-tuning and semi-supervised training with partially annotated ScanNet data provide the best overall results.}
    \resizebox{0.6\linewidth}{!}{
        \begin{tabular}{l|ccc}
            \toprule
            & \multicolumn{3}{c}{Acc@0.5IoU} \\ 
            \midrule
            & Unique & Multiple & Overall \\
            \midrule
            ScanRefer~\citep{chen2020scanrefer} & - & 12.17 & 12.17 \\
            3DVG-Trans~\citep{zhao20213dvg} & - & 14.22 & 14.22 \\
            \midrule
            Ours (w/o fine-tuning) & - & 19.64 & 19.64 \\
            Ours (w/ fine-tuning) & - & 24.41 & 24.41 \\
            Ours (w/ fine-tuning + extra) & - & \textbf{25.23} & \textbf{25.23} \\
            \bottomrule
        \end{tabular}
    }
    
    \label{tab:grounding_referit3d}
\end{table}

We compare our 3D visual grounding results against the baseline ScanRefer~\citep{chen2020scanrefer} and 3DVG-Transformer~\citep{zhao20213dvg} in Tab.~\ref{tab:grounding_referit3d}. As the descriptions in ReferIt3D dataset all refer to objects in the scene where multiple similar objects with the same class label are present, there is no such case that can be allocated to ``Unique'' subset where only one object with a specific class label can be found in the scene. Therefore, we allocate our results to ``Multiple'' and ``Overall''. Our method trained with the detector loss and the listener loss (marked ``Ours(w/o fine-tuning)'') clearly outperforms the baseline methods. Our results (marked ``Ours(w/ fine-tuning)'') are significantly improved after fine-tuning jointly with the speaker. Our best results are obtained after jointly training with speaker-listener architecture on partially annotated ScanNet data, as demonstrated in the last row in Tab.~\ref{tab:grounding_referit3d}.

\input{figures/captioning_referit3d}

\begin{table*}[ht!]
    \centering
    \caption{Comparison of the performance of our implementation of PointGroup using the Minkowski Engine against the original PointGroup (PG(*)) for instance segmentation.  We report the mAP for IoU threshold 0.5 on the ScanNet v2 validation set.  Our re-implementation using color gives comparable performance as the original PointGroup implementation.   Using multiview features, we are able to further improve the performance.}
    \resizebox{\textwidth}{!}{
        \begin{tabular}{l|c|cccccccccccccccccc}
            \toprule
            Method & mAP@0.5 & cab. & bed & chair & sofa & tab. & door & wind. & booksh. & pic. & cntr & desk & curt. & refrige. & s. curt. & toil. & sink & batht. & other \\
            \midrule
            PG (*) & 56.9 & 48.1 & 69.6 & \textbf{87.7} & 71.5 & 62.9 & 42.0 & 46.2 & 54.9 & 37.7 & 22.4 & 41.6 & 44.9 & 37.2 & 64.4 & 98.3 & \textbf{61.1} & 80.5 & 53.0 \\
            PG (Color) & 56.6 & 47.5 & 64.1 & 83.8 & \textbf{75.4} & 63.7 & 42.7 & 45.7 & 49.6 & \textbf{43.7} & 17.5 & 42.9 & \textbf{47.9} & 35.0 & 65.6 & \textbf{100.0} & 60.7 & \textbf{81.9} & 51.5 \\
            PG (Multiview) & \textbf{62.8} & \textbf{58.3} & \textbf{83.4} & 86.9 & 66.3 & \textbf{68.6} & \textbf{47.3} & \textbf{52.4} & \textbf{64.9} & 38.3 & \textbf{23.0} & \textbf{56.9} & 46.3 & \textbf{64.3} & \textbf{83.0} & 98.3 & 57.0 & 71.4 & \textbf{63.1} \\
            \bottomrule
        \end{tabular}
    }
    
    \label{tab:pointgroup_instance}
\end{table*}

\input{figures/grounding_referit3d}

\subsection{Qualitative Analysis}

\subsubsection{3D dense captioning}

We compare our results with object captions from Scan2Cap~\citep{chen2021scan2cap} in Fig.~\ref{fig:captioning_referit3d}. Object captions generated by Scan2Cap include more inaccurate spatial relationships. Also, those object captions cannot be used to uniquely localize the associated object. In contrast, our method produces more accurate and discriminative object captions with more spatial relationship information.

\subsubsection{3D visual grounding}

Fig.~\ref{fig:grounding_referit3d} compares our results with 3DVG-Transformer~\citep{zhao20213dvg} on ReferIt3D Nr3D dataset~\citep{achlioptas2020referit3d}. 3DVG-Transformer clearly suffers from overfitting issue, as it tends to predict that same object bounding box given different queries as inputs (see the third and fourth examples in the first row). Leveraging the speaker-listener architecture, our method can better distinguish object from the same class than 3DVG-Transformer.

\input{figures/extra}

\section{Additional Results on Extra ScanNet Data}
\label{sec:extra}

Fig.~\ref{fig:extra} showcase the intermediate dense captioning and visual grounding results for scans where no GT object captions are provided in the ScanRefer dataset~\citep{chen2020scanrefer}.
Those intermediate object captions and the matched object bounding boxes are used during the semi-supervised training of our speaker-listener architecture.
Our architecture produces plausible object captions with adequate and discriminative spatial relationships that inherently enables visual grounding.

\section{PointGroup Implementation Details}
\label{sec:pointgroup_details}

The official implementation of PointGroup uses SpConv~\citep{spconv}, a spatially sparse convolution library devoted to 3D data, to build its SparseConv-based U-Net architecture to encode point and cluster representation. We migrated the implementation of PointGroup from SpConv to MinkowskiEngine~\citep{choy20194d}, another auto-differentiation library for sparse tensors, since it outperformed SpConv by providing faster computation operations on GPU, user-friendly documentations and consistent code maintenance at the time the project was initiated. 

Following~\citet{jiang2020pointgroup}, we use the same hyperparameters for point voxelization and clustering. We set the maximum number of points per scene to $250,000$ by randomly adding small offsets to the point cloud and cropping out extra parts exceeding the predefined maximum scale of the scene if necessary.  Limiting the number of points to $250,000$ allows us to fit the model on a RTX 3090.  We augment each point cloud scene by jittering point coordinates slightly, mirroring about the YZ-plane, and rotation about the Z axis (up-axis)  randomly from 0 to $360^{\circ}$. We also apply elastic distortion, which was used by~\citet{jiang2020pointgroup}, to the scaled points. We share the same SparseConv-based U-Net architecture as~\citet{jiang2020pointgroup} for both backbone and ScoreNet except that the input data may contain mutiview features and normals instead of RGB colors. For each voxel, we encode the color, normal and multiview features extracted using ENet~\cite{dai20183dmv}, giving us a total input dimension of $134$.  To adapt PointGroup as an object detector, we obtain axis-aligned bounding boxes using predicted instance clusters by simply calculating their sizes and centers from points assigned to them. We set the thresholds of cluster scores as $0.09$ and the minimum cluster point number as $100$ to filter out bad cluster proposals. We train the PointGroup detector using Adam~\citep{kingma2014adam} with a learning rate of 2e-3, on the ScanNet train split with batch size 4 for 140k iterations until convergence.

\section{Detection and Segmentation Results}
\label{sec:pointgroup_vis}

\begin{table*}[ht!]
    \centering
    \caption{Comparison of object detection performance of PointGroup (PG) and VoteNet.   We report mAP with IoU threshold 0.5 on the ScanNet v2 validation set.  PointGroup produces more accurate bounding boxes than VoteNet, and using multiview features further improves performance over incorporating color as input directly.}
    \resizebox{\textwidth}{!}{
        \begin{tabular}{l|c|cccccccccccccccccc}
            \toprule
            Method & mAP@0.5 & cab. & bed & chair & sofa & tab. & door & wind. & booksh. & pic. & cntr & desk & curt. & refrige. & s. curt. & toil. & sink & batht. & other \\
            \midrule
            VoteNet & 33.5 & 8.1 & 76.1 & 67.2 & \textbf{68.8} & 42.4 & 15.3 & 6.4 & 28.0 & 1.3 & 9.5 & 37.5 & 11.6 & 27.8 & 10.0 & 86.5 & 16.8 & \textbf{78.9} & 11.7 \\
            PG (Color) & 44.6 & 25.2 & 69.1 & 77.1 & 67.3 &53.3 & 32.7 & 32.2 & 36.8 & 26.9 & 30.0 & 52.1 & \textbf{33.5} & 26.7 & 37.4 & 87.8 & 32.3 & 69.6 & 13.6 \\
            PG (Multiview) & \textbf{50.7} & \textbf{36.4} & \textbf{77.6} & \textbf{80.9} & 66.1 & \textbf{59.2} & \textbf{40.2} & \textbf{33.1} & \textbf{37.0} & \textbf{27.7} & \textbf{32.0} & \textbf{56.5} & 32.2 & \textbf{62.1} & \textbf{70.0} & \textbf{91.1} & \textbf{33.8} & 60.2 & \textbf{16.0} \\
            \bottomrule
        \end{tabular}
    }
    
    \label{tab:pointgroup_detection}
\end{table*}

\subsection{Quantitative results}

\paragraph{Instance segmentation.}

Tab.~\ref{tab:pointgroup_instance} compares the instance segmentation results of our PointGroup implementation against the original PointGroup (first row). With positions and colors as input, our implementation of PointGroup (second row) gives a similar performance as original PointGroup.  When replacing colors with multiview features and normals (last row), our PointGroup implementation significantly outperforms the original one. Our multiview-based PointGroup gives mAP@0.5 of $62.8$, which is close to the performance of the current state-of-the-art model HAIS~\cite{chen2021hierarchical}, which achieves $64.1$ on the validation set of ScanNet v2. Our implementation also surpasses the original PointGroup in training speed: given point coordinates and colors as input, it takes less than two days to train the model in our implementation, while the original one could take up to three days until convergence. 

\paragraph{Object Detection.}

We compare our object detection results before fine-tuning with the speaker-listener architecture against the VoteNet~\citep{qi2019deep} in Tab.~\ref{tab:pointgroup_detection}. Given positions and colors as input, our PointGroup detector (second row) clearly outperforms VoteNet. Using multiview features and normals instead of RGB colors, our PointGroup based detector gives improved detection results of 50.7 mAP@0.5, which outperforms the current state-of-the-art detectors~\citep{misra2021-3detr, zhang2020h3dnet} on the validation set of ScanNet v2 with gains of 3.7 and 2.6 respectively. Also, our PointGroup generates notably better detections for small and thin objects than VoteNet, such as picture (``pic.'') and counter (``cntr''). 

\subsection{Qualitative results}

\paragraph{Instance segmentation.}

We present our instance segmentation results in Fig.~\ref{fig:instance_comparsion}. Our PointGroup trained with multiview features and normals clearly generates better instance segmentation masks than our model with raw point colors as input, as it better segments out tiny objects leveraging the higher resolution of the multiview images. 

\paragraph{Object Detection.}

Fig.~\ref{fig:detection_comparsion} showcases the effectiveness of our PointGroup in object detection over VoteNet. Our PointGroup implementation produces much more accurate object bounding boxes due to the fine-grained per-point segmentation. Also, training with multiview normal features can further improve the quality of the generated bounding boxes in comparison with PointGroup trained with the raw point colors (the third column vs. the first column).

\input{figures/instance_comparsion}
\input{figures/detection_comparsion}

%% file: figures/captioning_referit3d.tex
\begin{figure*}[t]
    \centering
    \includegraphics[width=0.99\textwidth]{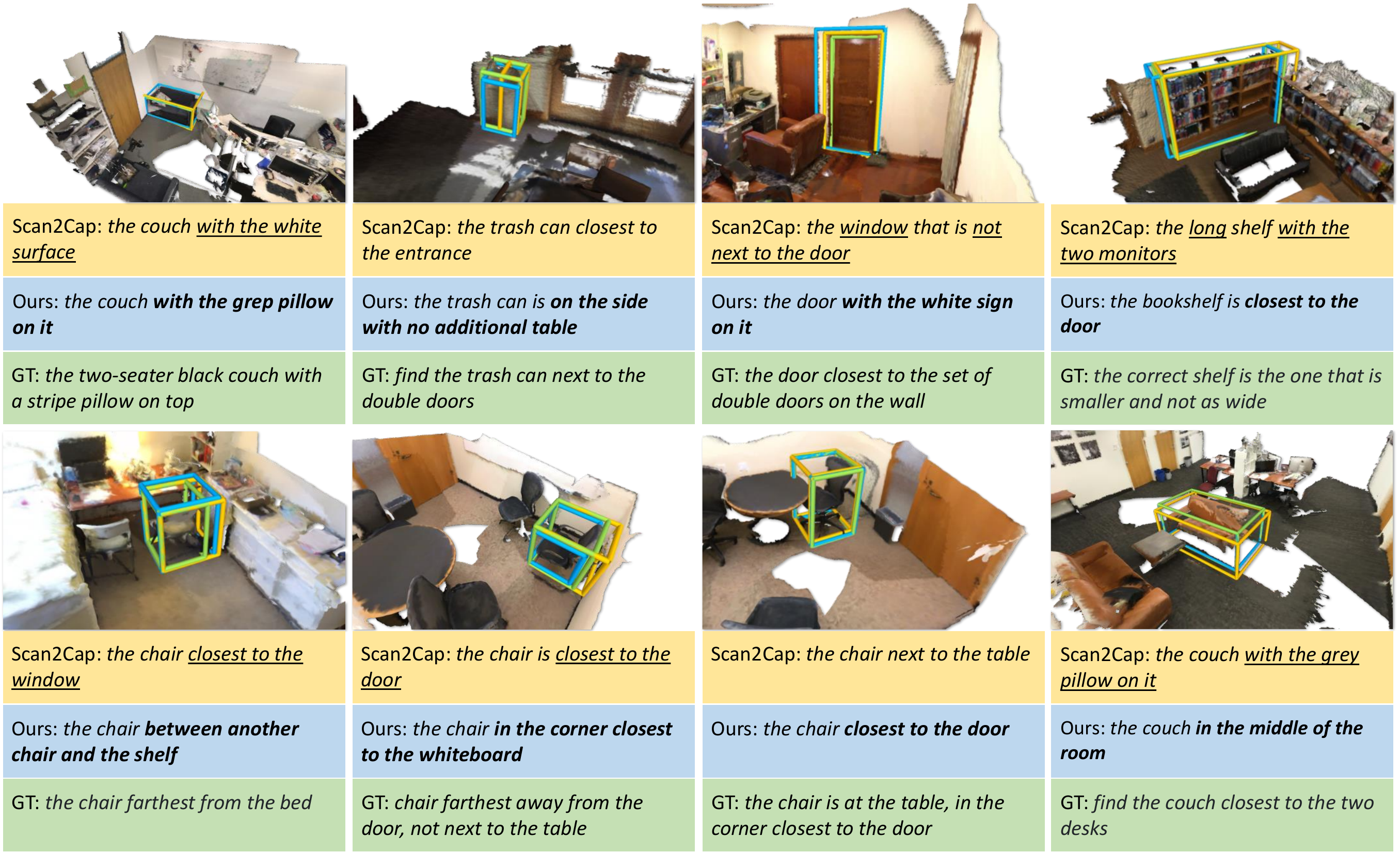}
    \caption{Qualitative results in 3D dense captioning task from Scan2Cap~\citep{chen2021scan2cap} and our method on ReferIt3D Nr3d dataset~\citep{achlioptas2020referit3d}. We underline the inaccurate words and mark the spatially discriminative phrases in bold.}
    \label{fig:captioning_referit3d}
\end{figure*}

%% file: figures/grounding_referit3d.tex
\begin{figure*}[t]
    \centering
    \includegraphics[width=0.99\textwidth]{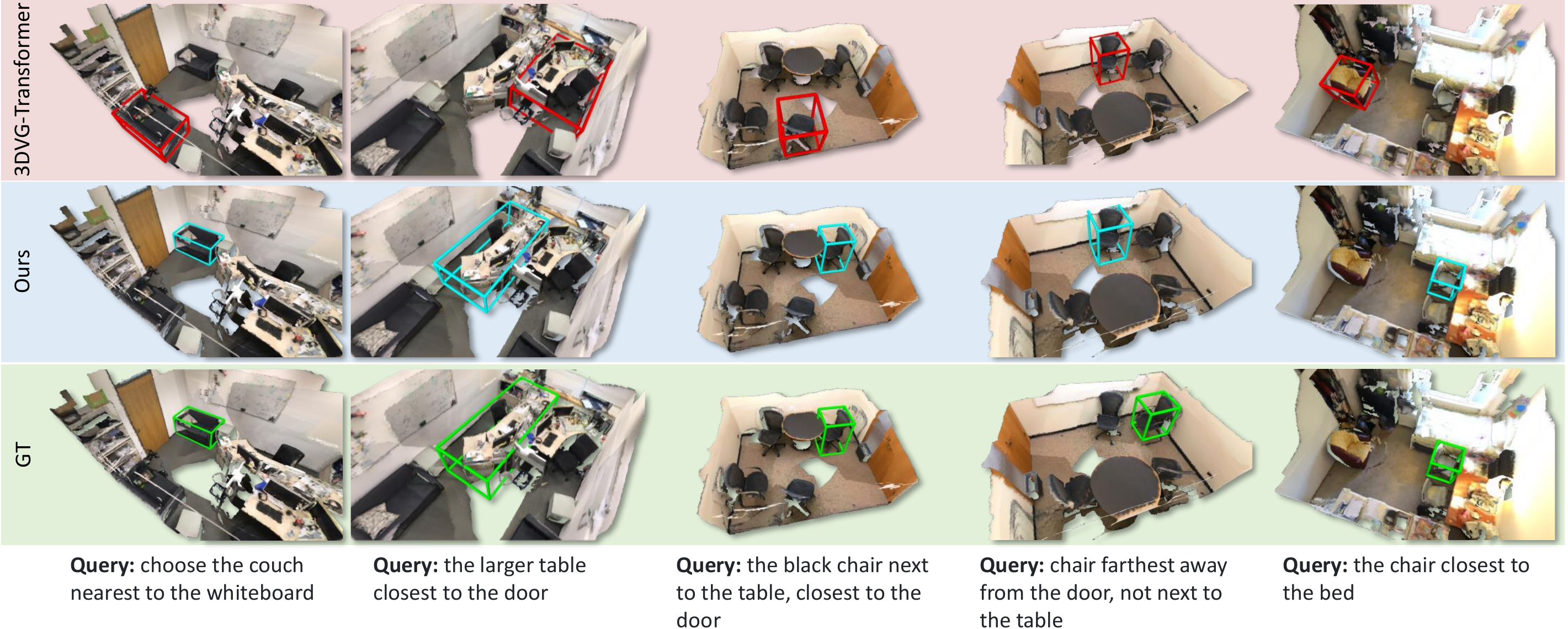}
    \caption{3D visual grounding results using 3DVG-Transformer~\citep{zhao20213dvg} and our method on ReferIt3D Nr3D dataset~\citep{achlioptas2020referit3d}.}
    \label{fig:grounding_referit3d}
\end{figure*}

%% file: figures/extra.tex
\begin{figure*}[t]
     \centering
     \begin{subfigure}{\textwidth}
         \centering
         \includegraphics[width=0.99\textwidth]{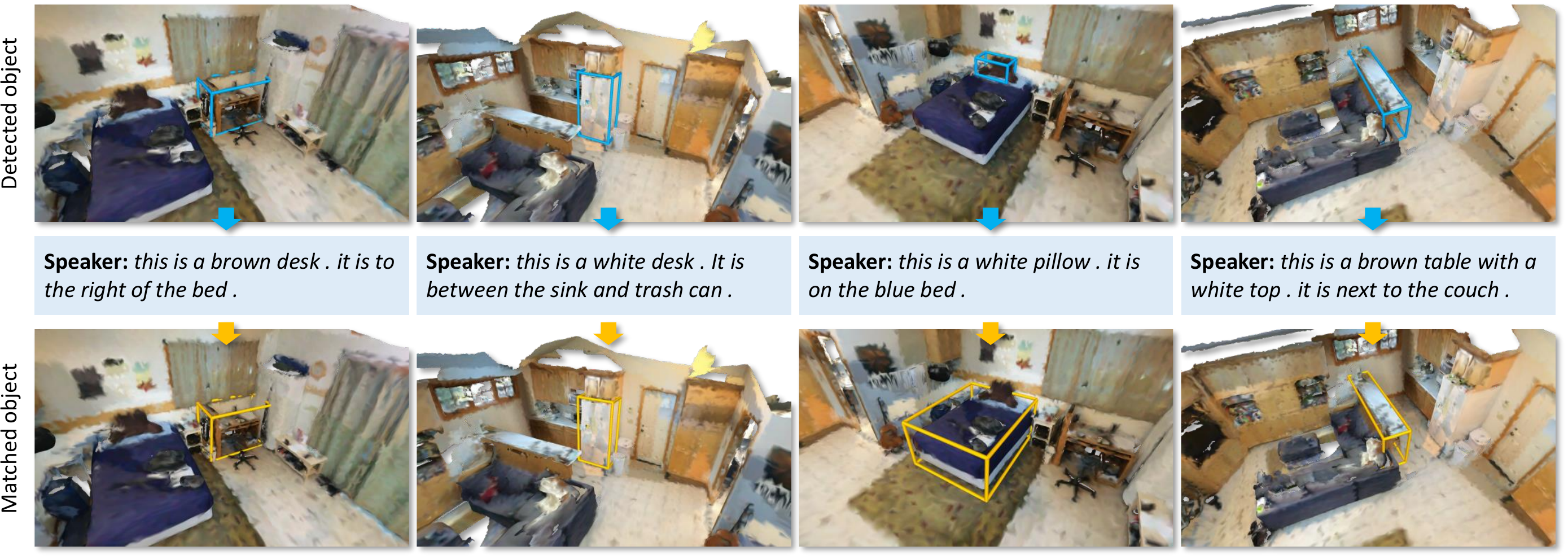}
         \caption{Scan scene0000\_01}
         \label{fig:extra_1}
     \end{subfigure}
     \hfill
     \begin{subfigure}{\textwidth}
         \centering
         \includegraphics[width=0.99\textwidth]{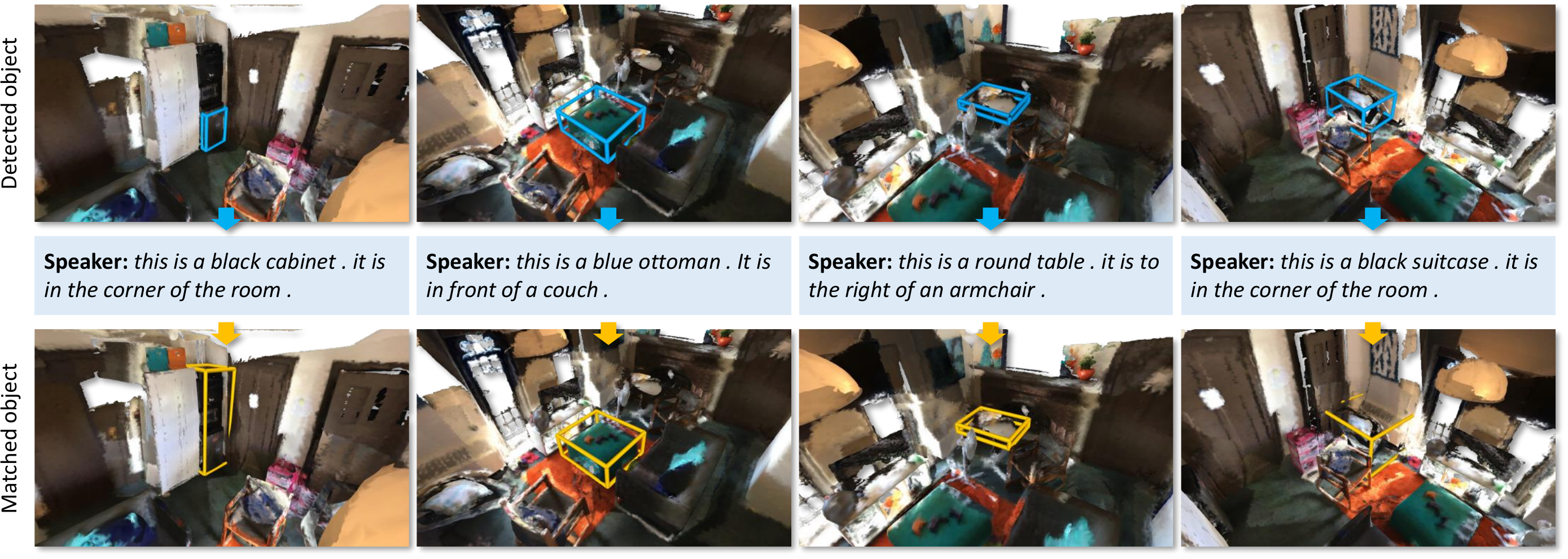}
         \caption{Scan scene0002\_01}
         \label{fig:extra_2}
     \end{subfigure}
     \caption{Intermediate dense captioning and visual grounding results in the Speaker-listener architecture for RGB-D scans where no GT object descriptions are provided in ScanRefer dataset~\citep{chen2020scanrefer}}
     \label{fig:extra}
\end{figure*}

%% file: figures/instance_comparsion.tex
\begin{figure*}[t]
    \centering
    \includegraphics[width=0.99\textwidth]{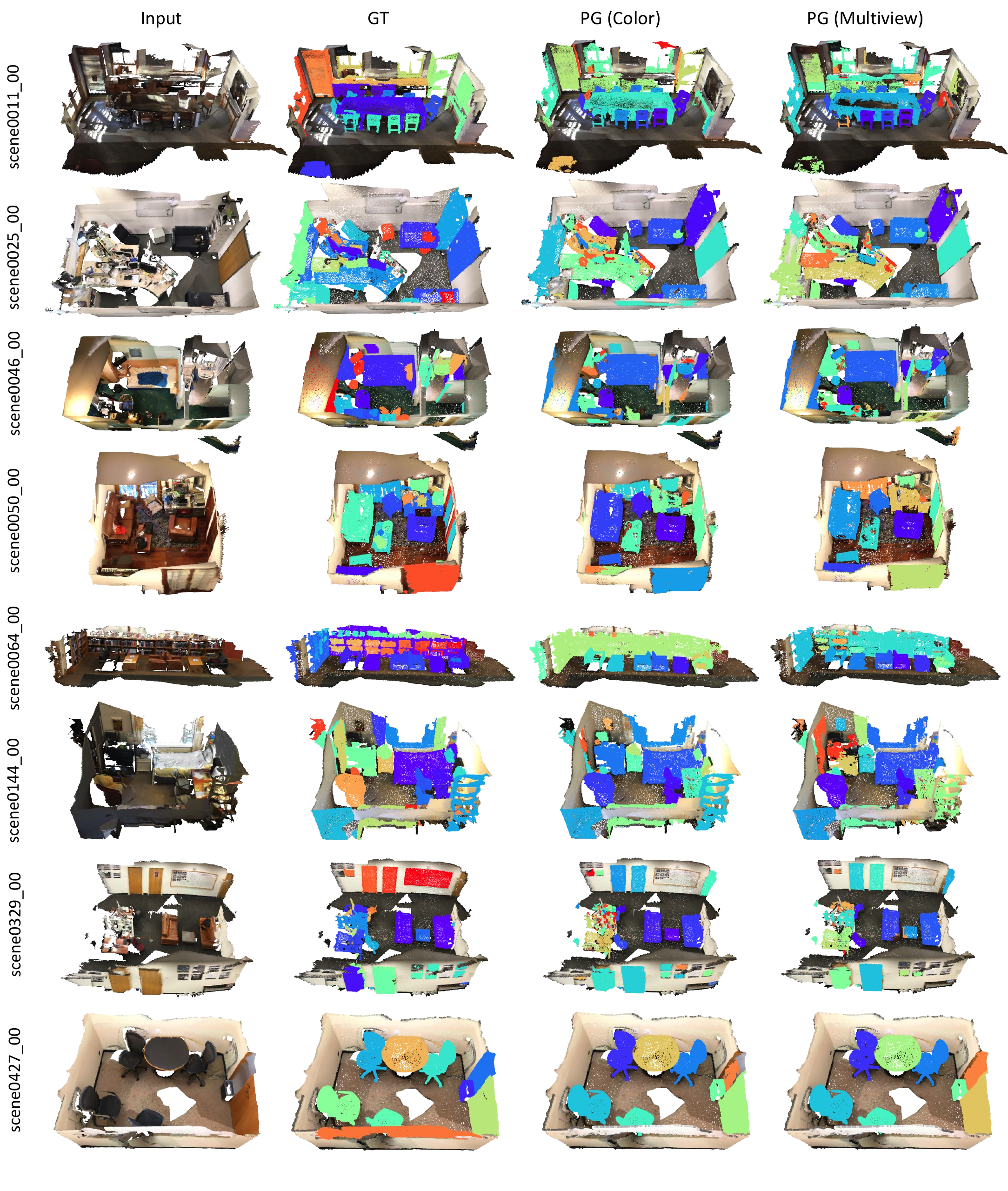}
    \caption{Qualitative results in instance segmentation task on the ScanNet v2 validation set.}
    \label{fig:instance_comparsion}
\end{figure*}

%% file: figures/detection_comparsion.tex
\begin{figure*}[t]
    \centering
    \includegraphics[width=0.99\textwidth]{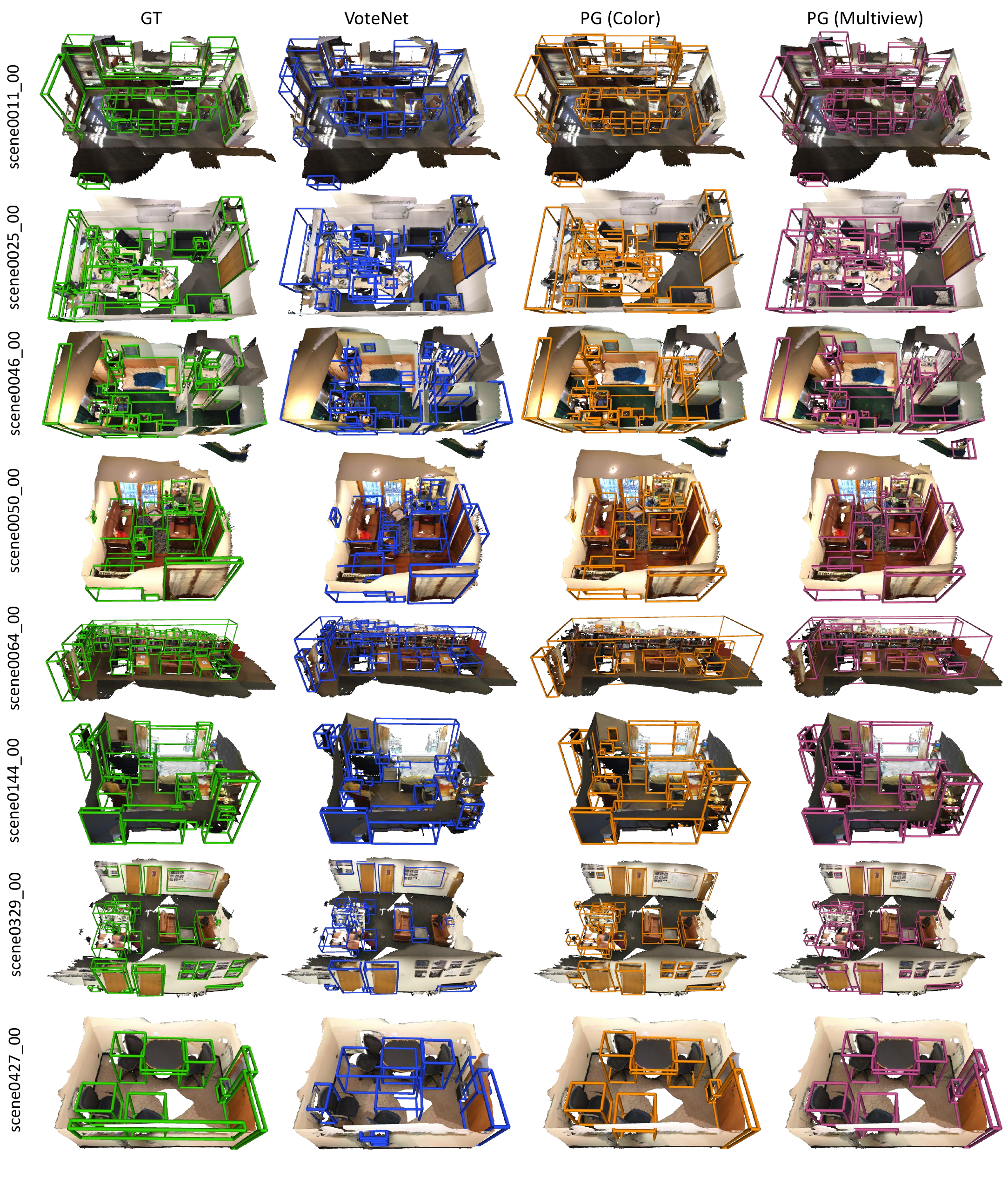}
    \caption{Qualitative results in object detection task on the ScanNet v2 validation set.}
    \label{fig:detection_comparsion}
\end{figure*}